\documentclass{article}

\usepackage{arxiv}

\usepackage[utf8]{inputenc}  
\usepackage[T1]{fontenc}     
\usepackage{amsmath,amssymb,amsfonts}
\usepackage{nicefrac}        
\usepackage{microtype}       

\usepackage{amsmath,amsfonts,bm,amssymb}

\def\eqref#1{equation~\ref{#1}}

\def\1{\bm{1}}

\DeclareMathAlphabet{\mathsfit}{\encodingdefault}{\sfdefault}{m}{sl}
\SetMathAlphabet{\mathsfit}{bold}{\encodingdefault}{\sfdefault}{bx}{n}

\usepackage{hyperref}        
\usepackage{url}             
\usepackage{booktabs}        
\usepackage{graphicx}
\usepackage{array}
\usepackage{xcolor}
\usepackage{xspace}
\usepackage{enumitem}
\usepackage{inconsolata}
\usepackage{framed}
\usepackage{listings}
\usepackage[round]{natbib}   
\usepackage{cleveref}        
\usepackage{authblk}         

\setlist{leftmargin=*, topsep=0.5em, itemsep=0pt, parsep=0pt}

\lstdefinelanguage{diff}{
  morecomment=[f][\color{teal}]{+},
  morecomment=[f][\color{red}]{-},
  morecomment=[f][\color{gray}]{@@},
  morecomment=[f][\color{gray}]{---},
  morecomment=[f][\color{gray}]{+++},
}
\lstdefinestyle{verus-example}{
  basicstyle=\small\ttfamily,
  keywordstyle=\bfseries,
  morekeywords={fn,ensures,return,let,assert,exists},
  columns=fullflexible,
  keepspaces=true,
  showstringspaces=false,
  frame=single,
  framerule=0.3pt,
  aboveskip=4pt,
  belowskip=0pt
}

\crefname{figure}{Fig.}{Figs.}
\Crefname{figure}{Fig.}{Figs.}
\crefname{section}{Sec.}{Secs.}
\Crefname{section}{Sec.}{Secs.}
\crefname{equation}{Eq.}{Eqs.}
\Crefname{equation}{Eq.}{Eqs.}
\crefname{appendix}{App.}{App.}
\Crefname{appendix}{App.}{App.}
\crefname{table}{Tab.}{Tabs.}
\Crefname{table}{Tab.}{Tabs.}

\newcommand{\sys}{\textsc{CodeWalk}\xspace}

\newif\ifcomments
\commentsfalse
\ifcomments
  \providecommand{\sahil}[1]{{\protect\color{red}{\bf [sahil: #1]}}}
  \providecommand{\alvin}[1]{{\protect\color{purple}{\bf [alvin: #1]}}}
  \providecommand{\jiaxin}[1]{{\protect\color{teal}{\bf [jiaxin: #1]}}}
  \providecommand{\runyuan}[1]{{\protect\color{blue}{\bf [runyuan: #1]}}}
  \providecommand{\neel}[1]{{\protect\color{green}{\bf [neel: #1]}}}
\else
  \providecommand{\sahil}[1]{}
  \providecommand{\alvin}[1]{}
  \providecommand{\jiaxin}[1]{}
  \providecommand{\runyuan}[1]{}
  \providecommand{\neel}[1]{}
\fi

\title{CodeWalk: Generating Coding Benchmarks by Walking a Problem Graph}

\author[1]{Jiaxin Fang}
\author[1]{Runyuan He}
\author[1]{Arnav Gupta}
\author[1]{Sahil Kumawat}
\author[1]{Huanzhi Mao}
\author[1]{Alvin Cheung}
\affil[1]{University of California, Berkeley \\ \texttt{\{jiaxinfang, runyuan, 12arnavg, sahilk\_17, huanzhimao\}@berkeley.edu, akcheung@cs.berkeley.edu}}

\renewcommand{\shorttitle}{CodeWalk: Generating Coding Benchmarks by Walking a Problem Graph}

\hypersetup{
  pdftitle={CodeWalk: Generating Coding Benchmarks by Walking a Problem Graph},
  pdfauthor={Jiaxin Fang, Runyuan He, Arnav Gupta, Sahil Kumawat, Huanzhi Mao, Alvin Cheung},
  pdfkeywords={code generation, benchmarks, compositional generalization, LLM evaluation},
}

\begin{document}
\maketitle

\begin{abstract}
High-quality coding problems are expensive to collect. Benchmarks such as SWE-bench rely on substantial human curation, while synthetic alternatives typically rely on costly model generation and validation. In this work, we show that new coding problems can be created at almost no cost by composing existing ones. We formulate problem generation as code composition: verified atomic problems are combined into level-$N$ tasks, and a task is kept only if every atomic problem remains unsolved and a joint reference solution passes the same check used at evaluation time. The entire task-construction process is programmatic and requires no language-model generation or validation. We instantiate this approach in four coding settings: multi-bug repair, proof completion, complexity analysis, and clone equivalence. The generated tasks are challenging for frontier models, which all degrade as compositional depth increases. The generated tasks are also useful as training data: rejection-sampling fine-tuning on them improves Qwen3.5-9B at every composition level.
\end{abstract}


\section{Introduction}
\label{sec:intro}

High-quality coding problems are expensive to collect. Benchmarks such as SWE-bench~\citep{jimenez2024swebench} are assembled by mining repositories and curating the results manually, and the effort of making each instance reliable is large enough that follow-up work has been devoted to keeping the pool fresh as repositories move on~\citep{badertdinov2025swerebench,
badertdinov2026swerebenchv2}. Synthetic pipelines replace curation cost with inference cost: a language model proposes problems, and in most designs another model or an execution loop validates them~\citep{yang2025swesmith, pan2024swegym, jain2025r2egym}. This trades one expense for another, and it bounds the difficulty of what is produced by what the generator can produce, while the labels are only as trustworthy as the process that assigned them. Meanwhile the problems these pipelines produce are being solved: function-level benchmarks such as HumanEval and MBPP are effectively saturated, with frontier models above $90\%$, and the benchmarks themselves are no longer able to separate strong models~\citep{dai2024mhpp, jain2024livecodebench}.


Cost, however, is not the only reason to look past collection. Even an unlimited supply of curated problems would leave a gap, because real work rarely presents one defect, one lemma, or one function in isolation: problems arrive together and interact. Handling them jointly is harder than handling each alone, and the added difficulty is not merely more work of the same kind. It
comes from requirements that belong to no single problem, such as edits that interfere with one another or intermediate results that must satisfy what consumes them. This gap is well documented outside software engineering: models that answer each sub-question correctly still fail on the question that combines
them~\citep{press2023measuring}, accuracy falls as composed steps
accumulate~\citep{dziri2023faith}, and merely lengthening a problem makes it harder~\citep{mirzadeh2024gsmsymbolic}. Coding benchmarks have begun to probe this by chaining tasks~\citep{jin2026chainswe}, but remain overwhelmingly
atomic. Composing existing problems is therefore not only cheap: it produces difficulty that collecting more atomic problems cannot.

In this paper, we show that new coding problems can be created at {\em almost no cost by strategically composing existing ones}, and composition can be modeled as a graph: the nodes are verified atomic problems that carry executable verifiers; an edge indicates that two problems can be composed.
A ``level-$N$'' task is then a group of $N$ nodes collected by walking this graph and placed into a single combined program. We admit the new task only if two checks pass: none of the $N$ problems is already solved in the combined program, so the task really is at level $N$; and the combined reference solution passes the same tests or verifier we later use to grade models, so the task is known to be solvable. The entire process is programmatic: the graph, the walk, the composition, and admission checks are deterministic programs over human-authored material, without the need for any language model. Difficulty is then a parameter of the generator: raising $N$ simply means walking further on the graph, so harder tasks come from
problems we already have rather than from new mining or annotation, and the largest level we can build is limited by how many problems can be linked, not by how strong a generator model is. Composing existing problems to control difficulty has been used to build evaluation sets that resist contamination~\citep{fang2026playingpsychic}; we use it as
the construction principle for benchmarks with executable and formal correctness checks.

We implement \sys (\cref{sec:formulation}) and use it to generate coding problems in four settings, chosen to exercise different parts of the formulation (\cref{sec:approach}). Evaluating the generated tasks on frontier models, we find that accuracy degrades significantly as compositional depth increases, matching what manually curated harder benchmarks achieve~\citep{dai2024mhpp, jin2026chainswe} at a fraction of the cost. The sharpest drop shows that the strongest model falls from $94\%$ at $N = 1$ to $35\%$ at $N = 3$ (\cref{sec:experiments}). The decline is steeper than independent single-problem failure would predict, and it is composition rather than the difficulty of the individual problems that breaks the strong models. We also study agentic RFT on the generated clone-equivalence and multi-bug repair tasks: Qwen3.5-9B
attempts each task through multi-step tool interactions, task verifiers select successful trajectories for fine-tuning, and the updated agent collects new trajectories as the training pool expands to higher composition levels (\cref{sec:rft}).

In sum, we make the following contributions in this paper:
\vspace{-0.1in}
\begin{itemize}
  \item \textbf{Problem generation as code composition.} We create new coding problems by composing verified existing ones, at compute cost rather than curation or inference cost, and admit a task only if every atomic problem remains unsolved and a joint reference passes the evaluation-time check.
  \item \textbf{Composition as graph walks.} We formulate new problem composition as a graph walk, where atomic problems are nodes, a domain's composition relation defines the edges, and the level-$N$ is the length of the walk that forms a task. Adding a new domain only requires a new edge relation. 
  %
  \item \textbf{Evaluation using frontier models.} We generate over 1700 tasks across four coding domains without a single language-model call; the most expensive domain takes around four hours of compute. Accuracy declines with $N$ for every frontier model, and the drop exceeds what independent per-problem failure predicts.
  %
  \item \textbf{Agentic rejection-sampling fine-tuning.} We use generated tasks as interactive environments for Qwen3.5-9B to collect tool-use trajectories, retain those that pass task verifiers, and fine-tune the agent on the successful trajectories. We evaluate task success as this collection-and-training cycle incorporates progressively more complex tasks.
\end{itemize}
\section{\sys: Model-Free Data Generation}
\label{sec:formulation}

\begin{figure*}[t]
  \centering
  \includegraphics[width=\textwidth]{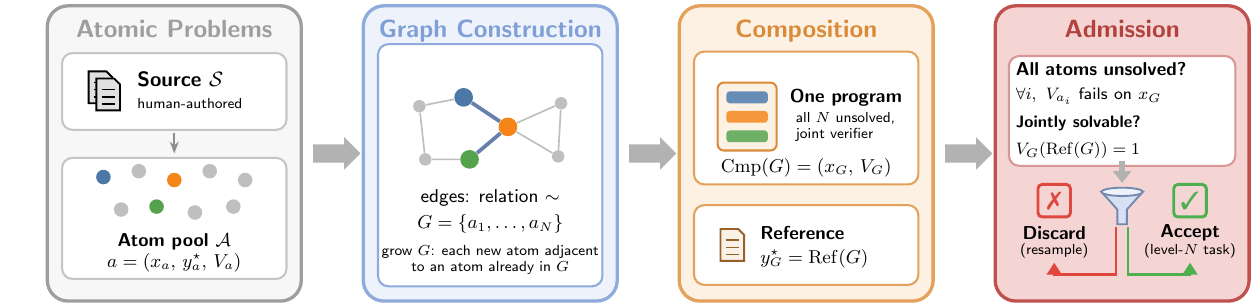}
  \caption{Task generation pipeline. Atoms are extracted from a source; $N$ interacting atoms grown on the problem graph are composed into a single task $(x_G, V_G)$ with reference $\mathrm{Ref}(G)$. Groups that fail admission are discarded and resampled.}
  \label{fig:pipeline}
  \vspace{-0.3in}
\end{figure*}






We generate coding tasks that must be solved as a whole rather than one problem at a time (\cref{fig:pipeline}). We call $N$ the \emph{level} of a task, the number of atomic problems it combines, and we generate a level-$N$ task by walking a graph whose nodes are existing atomic problems and whose edges indicate which problems can be composed.

Every step of this generator is a deterministic program over existing, human-authored problems. No language model selects, assembles, labels, or validates a task. Generation therefore costs compute rather than inference budget; reference solutions are not model-produced and are kept only after passing the same check used to grade models (\cref{sec:admission}); and difficulty is not bounded by what a generation model can produce, so $N$ can be
raised as far as the graph allows. The rest of this section describes the generator, and \cref{sec:approach} gives its per-domain specializations.

\vspace{-0.06in}
\subsection{Atomic problems}
\label{sec:atoms}
\vspace{-0.06in}
A domain supplies a source $\mathcal{S}$ of raw tasks together with an extraction procedure yielding a pool $\mathcal{A}$ of atomic problems, or atoms for short.
An atom is a tuple $a = (x_a,\ y^\star_a,\ V_a)$,
where $x_a$ is a problem context, $y^\star_a$ a reference solution, and $V_a:\mathcal{Y}\to\{0,1\}$ an executable verifier deciding the correctness of a candidate output. An atom constitutes a level-$1$ task. The output space $\mathcal{Y}$ is domain-specific: it may consist of source edits, program annotations, or predicted labels.

\vspace{-0.06in}
\subsection{Graph construction}
\label{sec:graph}
\vspace{-0.06in}
A level-$N$ task is obtained by placing $N$ interacting atoms into a single state that must be solved as a whole. We select such groups by walking a graph over the pool $\mathcal{A}$.
\vspace{-0.12in}
\paragraph{Interaction.}
A relation ${\sim}$ over $\mathcal{A}$ specifies which pairs of atomic problems can be combined to form a higher-level task, and applying it to all pairs induces the \emph{problem graph}, whose nodes are the atomic problems and whose edges are the pairs $a_i \sim a_j$. What an edge means is domain-specific. In multi-bug repair, $a_i \sim a_j$ when the two bug fixes modify the same file, so that an edit for one can interfere with the other; the relation is symmetric. In proof completion, complexity analysis, and clone equivalence, $a_i \sim a_j$ when the output of $a_i$ has the type $a_j$ consumes; the relation is then directed.

\vspace{-0.12in}
\paragraph{Sampling.}
Given the problem graph, we build a level-$N$ task by picking a node $a_1$ and following edges $N-1$ times: at each step we pick, uniformly at random, an unvisited node adjacent to any node already picked, giving $G=\{a_1,\dots,a_N\}$. If no such node remains, we start over. Because each new node is adjacent to an earlier one, every atomic problem in $G$ interacts with at least one other: in multi-bug repair the $N$ bugs are linked by shared files (\cref{sec:repair}), and under a directed relation $G$ is a chain in which each atomic problem feeds the next (\cref{sec:verus}). Building $G$ this way costs $O(N)$ and extends to any level the graph's connectivity supports.

\vspace{-0.06in}
\subsection{Composition}
\label{sec:composed}
\vspace{-0.06in}

\paragraph{Composed task.}
An operator $\mathrm{Cmp}$ turns $G$ into a single task, $\mathrm{Cmp}(G) = (x_G,\ V_G)$, where $x_G$ is one program containing all $N$ atomic problems, each still unsolved, and $V_G$ is the verifier for the combined task. In multi-bug repair $\mathrm{Cmp}$ has to find a commit at which all $N$ bugs are present (\cref{sec:repair}); in the other three domains $\mathrm{Cmp}$ builds it by chaining the atomic problems (\cref{sec:verus}).

\vspace{-0.12in}
\paragraph{Reference solution.}

A map $\mathrm{Ref}$ gives the reference solution $y^\star_G = \mathrm{Ref}(G)$, which need not be the union of the atomic reference solutions: in complexity analysis it is a single label
derived from the atomic labels rather than a concatenation of them
(\cref{sec:complexity}).

\vspace{-0.06in}
\subsection{Admission}
\label{sec:admission}
\vspace{-0.06in}
A sampled $G$ becomes a benchmark task only if it is neither partly pre-solved nor unsolvable. We check both against the same verifiers used to grade models:
\begin{align}
  &\textbf{(1) All atomic problems unsolved:}\quad
    \forall i,\ V_{a_i} \text{ fails on } x_G,
    \label{eq:live}\\[2pt]
  &\textbf{(2) Jointly solvable:}\quad
    V_G\big(\mathrm{Ref}(G)\big) = 1 .
    \label{eq:solv}
\end{align}
\Cref{eq:live} requires every atomic problem to be unsolved in the combined program; otherwise composition may leave one already satisfied, or absent altogether, and the task is really below level $N$. \Cref{eq:solv} requires the reference solution to pass $V_G$, so that a model's failure reflects difficulty rather than an ill-posed task. Groups failing either check are discarded and we sample another walk.
 



\section{Using \sys to Different Coding Domains}
\label{sec:approach}


We have used \sys to generate problems in four coding domains, each chosen to exercise different parts of the formulation. Every domain defines its source $\mathcal{S}$, relation ${\sim}$, composition $\mathrm{Cmp}$, reference $\mathrm{Ref}$, and verifier $V_G$. \Cref{tab:instantiations} summarizes them. Admission is decided by running tests, by a theorem verifier, or by type checking and execution, and no step invokes a language model.

\begin{table}[t]
\centering
\footnotesize
\hyphenpenalty=10000\exhyphenpenalty=10000
\setlength{\tabcolsep}{3pt}
\renewcommand{\arraystretch}{1.05}
\newcolumntype{L}[1]{>{\raggedright\arraybackslash}p{#1}}
\begin{tabular}{@{}lL{2.55cm}L{2.55cm}L{2.55cm}L{2.55cm}@{}}
\toprule
 & \textbf{Multi-bug repair} & \textbf{Proof completion} & \textbf{Complexity} & \textbf{Clone equivalence} \\
\midrule
Source $\mathcal{S}$ & repository with commit history & Verus-Bench programs & CodeComplex snippets & labeled Java method pairs \\
Atom $a$ & $(c_b,g_b,t_b,F_b)$ & $(f,S_a)$ & $(s,f,\tau_{\mathrm{in}},\tau_{\mathrm{out}},\ell)$ & $(p_1,p_2,T,\lambda)$ \\
Interaction ${\sim}$ & patches share a file & \resizebox{\linewidth}{!}{$\tau_{\mathrm{out}}(a_i)\preceq\tau_{\mathrm{in},k}(a_j)$} & $\tau_{\mathrm{out}}(c_i)\preceq\tau_{\mathrm{in}}(c_j)$ & $\tau_{\mathrm{out}}(e_i)=\tau_{\mathrm{in}}(e_j)$ \\
Composition $\mathrm{Cmp}$ & locate $c_\star$, apply test patches & chain under driver; remove annotations & chain under driver & chain each side under driver \\
Reference $\mathrm{Ref}$ & union of gold patches & annotations $y_G^\star$ & label derived from atomic labels & \textsc{eq} unless one \textsc{neq} atom \\
Verifier $V_G$ & $\mathcal{F}$ passes & Verus accepts restored program & predicted label matches & composites match label \\
Checked by & native test runner & Verus & type check and derivation & compile and execute \\
\bottomrule
\end{tabular}
\caption{Applying \sys to four coding domains. Each column gives one domain's source, relation, composition, reference and verifier.}
\label{tab:instantiations}
\vspace{-0.15in}
\end{table}

\vspace{-0.06in}
\subsection{Co-located multi-bug repair}
\label{sec:repair}

\vspace{-0.06in}
\paragraph{Atoms.}
$\mathcal{S}$ is a code repository with commit history, from which we extract bug instances. An atom is a bug instance $b = (c_b,\ g_b,\ t_b,\ F_b)$,
where $c_b$ is a commit at which the bug is present, $g_b$ is its gold source patch, $t_b$ is a test patch introducing the tests that exercise it, and $F_b$ are the tests that fail before the fix and should pass after it. The verifier $V_b$ runs $F_b$ and checks that all tests pass.

\vspace{-0.06in}
\paragraph{Notation.}
We define $\mathrm{apply}(c,P)$ as the code tree obtained by checking out commit $c$ and applying a set of patches $P$, and
$\mathrm{run}(W,F)\in\{\textsc{pass},\textsc{fail}\}$ as the outcome of executing test set $F$ on code tree $W$. All execution uses the repository's native test runner inside a container image pinned to its environment. We write $d(b)$ for the date of $b$'s fixing commit.

\vspace{-0.06in}
\paragraph{Interaction.}
We write $\mathrm{files}(g)$ for the files a patch modifies, recovered from the diff's headers. Two bugs are related when their gold patches share a file:
$
  b_i \sim b_j \iff \mathrm{files}(g_i)\cap\mathrm{files}(g_j)\neq\varnothing .
$
The relation is symmetric, as co-located edits contend for the same source. Every bug in a group therefore shares a file with another bug in the group, though the $N$ bugs need not all share one file.

\vspace{-0.06in}
\paragraph{Composition.}
A commit at which all $N$ bugs are present may already exist, so $\mathrm{Cmp}$ only has to find it: we order the group by fixing date, $d(b_1)\le\cdots\le d(b_N)$, and take the earliest commit $c_\star := c_{b_1}$, which precedes every fix in $G$ and where all $N$ bugs are therefore unfixed.
We write $t_i$ and $F_i$ for the test patch and target tests of $b_i$, and let $T=\{t_1,\dots,t_N\}$ and $\mathcal{F}=\bigcup_i F_i$. The composed task is the code tree $x_G \;=\; \mathrm{apply}(c_\star,\, T)$,
presented together with $\mathcal{F}$. A solution is a source edit $\delta$ restricted to non-test files, and
$V_G(\delta) \;=\; \mathbb{1}\!\left[\mathrm{run}(x_G \oplus \delta,\ \mathcal{F})=\textsc{pass}\right]$,
where $x_G \oplus \delta$ applies the edit to $x_G$. The reference solution is the union of gold patches, i.e., $\mathrm{Ref}(G)=\{g_1,\dots,g_N\}$.

\begin{figure}[t]
  \centering
  \includegraphics[width=\linewidth]{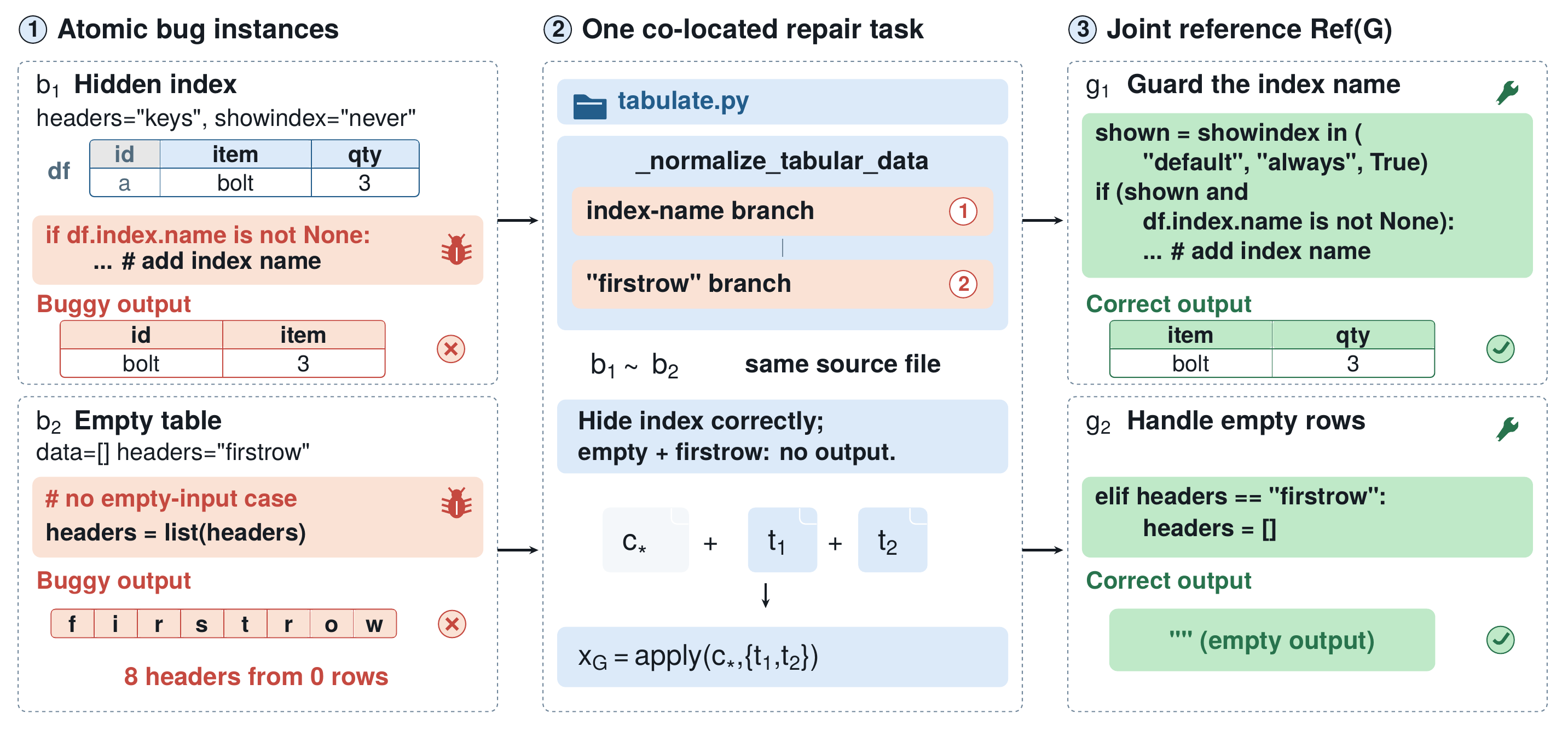}
  \caption{A level-$2$ repair task in \texttt{python-tabulate}, with simplified code and inputs. Hiding a named index shifts column headers, while empty input with \texttt{headers="firstrow"} creates spurious headers. Both bugs coexist at $c_\star$; their test patches form one task, and the reference combines the two source fixes.}
  \label{fig:bugfixing-example}
\vspace{-0.2in}
\end{figure}

\vspace{-0.06in}
\paragraph{Admission.}
In this domain, both checks reduce to test execution:
\begin{align*}
  &\textbf{(1)}\quad \forall i,\;
    \mathrm{run}\!\big(\mathrm{apply}(c_\star,T),\,F_i\big)=\textsc{fail},\\
  &\textbf{(2)}\quad
    \mathrm{run}\!\big(\mathrm{apply}(c_\star,\,T\cup\{g_1,\dots,g_N\}),\,\mathcal{F}\big)=\textsc{pass}.
\end{align*}
Date ordering makes co-location plausible; (1) confirms it empirically, rejecting groups in which a later bug is not yet expressible at $c_\star$ or its test patch fails to apply, while (2) certifies that a correct joint repair exists.

\vspace{-0.06in}
\paragraph{Construction.}
Bug instances are drawn from SWE-rebench~\citep{badertdinov2025swerebench} and
SWE-rebench-V2~\citep{badertdinov2026swerebenchv2}, which mine issue-resolving tasks from live GitHub repositories and ship each instance with a pre-built container image. Within each repository we build the graph over its instances, grow groups as described in~\cref{sec:graph}, and admit each candidate by execution. \Cref{fig:bugfixing-example} illustrates a level-$2$ task in \texttt{python-tabulate}, where two bugs in the same file must be repaired together.
The benchmark contains 200 tasks at $N{=}1,2$ and 168 at $N{=}3$,
spanning 270 distinct repositories. \Cref{app:bug-construction} describes the construction details and admission yields.

\vspace{-0.06in}
\subsection{Proof completion}
\label{sec:verus}

\vspace{-0.06in}
\paragraph{Atoms.}
$\mathcal{S}$ contains verified Rust functions. An atom is
$a=(f,S_a)$, where $f$ includes its precondition and supporting definitions.
We express its original postcondition as a fixed Boolean predicate
$S_a(x,r)$, which states what result $r$ must satisfy for inputs $x$.
The reference annotations $y_a^\star$ prove this specification in Verus,
so $V_a(y_a^\star)=1$.

\vspace{-0.04in}
\paragraph{Interaction.}
Let $\tau_{\mathrm{out}}(a_i)$ be the return type of $a_i$ and
$\tau_{\mathrm{in},k}(a_j)$ the type of argument $k$ of $a_j$.
We connect functions from different source programs when some argument
$k$ satisfies
\[
  \tau_{\mathrm{out}}(a_i)\preceq\tau_{\mathrm{in},k}(a_j).
\]
The relation $\preceq$ means the output can be passed to that input,
directly or through an allowed conversion, such as using a vector as a
slice. This defines the directed edge $a_i\sim a_j$; remaining arguments
become separate scalar inputs. A level-$N$ task is a chain
$G=(a_1,\ldots,a_N)$, excluding auxiliary functions from the count.

\vspace{-0.04in}
\paragraph{Composition.}
For each $a_i$, write $f_i$ for its function and $S_i:=S_{a_i}$ for its
postcondition, and let $x_0$ be the initial input.
$\mathrm{Cmp}(G)=(x_G,V_G)$ renames functions to avoid collisions,
chains their calls as $x_i=f_i(x_{i-1})$ for $i=1,\ldots,N$,
and generates a driver: a wrapper that executes these calls in order
and returns $x_N$. Additional task inputs and allowed conversions are
omitted from this notation.
For proof completion, the driver, \texttt{chain\_driver}, also asserts
$\bigwedge_{i=1}^{N}S_i(x_{i-1},x_i)$.
The task $x_G$ removes component postconditions and proof annotations,
including loop invariants and proof blocks, while keeping the code,
preconditions, $S_i$ definitions, and driver.
The reference $\mathrm{Ref}(G)=y_G^\star$ supplies the original annotations.
For a candidate solution $y$, $V_G(y)=1$ only when the completed program
verifies with the original $S_i$ and driver restored; missing blocks and
proof-bypass constructs are rejected.

\vspace{-0.04in}
\paragraph{Admission.}
We require $V_G(y_G^\star)=1$ and that the uncompleted program $x_G$
is well formed but fails verification. The first check establishes
a reference solution; the second ensures that proof work remains
at the whole-program level.

\vspace{-0.04in}
\paragraph{Construction.}
We use verified reference programs from
Verus-Bench~\citep{yang2025autoverus}. We form chains by following compatible function connections and apply
both admission checks to each complete program. The benchmark contains 100 tasks at each of
$N\in\{1,2,3,10\}$. \cref{app:verus-construction} describes
the construction details and reference-verification yields.

\vspace{-0.06in}

\subsection{Code complexity analysis}
\label{sec:complexity}

\vspace{-0.06in}
\paragraph{Atoms.}
$\mathcal{S}$ is CodeComplex~\citep{baik2025codecomplex}, a corpus of Python programs annotated across asymptotic time-complexity classes. An atom is
$c=(s,f,\tau_{\mathrm{in}},\tau_{\mathrm{out}},\ell)$, where
$f:\tau_{\mathrm{in}}\to\tau_{\mathrm{out}}$ is the designated entry function
and $\ell$ its annotated class. Auxiliary functions remain embedded but are not
exposed as composition boundaries. $V_a$ checks agreement with $\ell$.

\paragraph{Interaction.}
The relation is type compatibility as in~\cref{sec:verus}, here $\tau_{\mathrm{out}}(c_i) \preceq \tau_{\mathrm{in}}(c_j)$ with $\preceq$ being the language's compatibility relation, requiring that neither atom's semantics change.

\vspace{-0.06in}
\paragraph{Composition.}
For $G=(c_1,\ldots,c_N)$, we use the same renaming, chaining, and driver
construction as in~\cref{sec:verus}. The task $x_G$ is the resulting program,
and the model predicts its end-to-end complexity class.
The reference $\mathrm{Ref}(G)$ is derived from the atomic labels $\ell_i$
and the input- and output-size growth along the chain; $V_G$ checks the
predicted label against this reference.


\vspace{-0.06in}
\paragraph{Admission.}
(2) holds by validation of the derivation. Lacking a per-atom verifier, (1) is enforced in the form $\mathrm{Ref}(G)\neq\max_i \ell_i$: a chain whose answer coincides with its dominant component is solvable without composing, so it plays
the role of a pre-solved task and is discarded.

\vspace{-0.06in}
\paragraph{Construction.}

Atoms are drawn from CodeComplex~\citep{baik2025codecomplex}, from which we keep one snippet per problem whose parameter and return types are well defined, giving a pool of 277 atoms. Chains of length $N \in \{1,\dots,10\}$ are grown following~\cref{sec:graph} and admitted by the derivation check. The benchmark contains 80 tasks at each of $N \in \{1,2,3,10\}$.
\Cref{app:complex-construction} describes the construction details and admission yields.

\vspace{-0.06in}
\subsection{Clone Equivalence}
\label{sec:clone}

\vspace{-0.06in}
\paragraph{Atoms.}
$\mathcal{S}$ is a database of human-labeled Java method pairs. An atom is an equivalence instance $e = (p_1,\ p_2,\ T,\ \lambda)$,
where $p_1,p_2$ are method snippets, $T$ a test suite, and
$\lambda\in\{\textsc{eq},\textsc{neq}\}$ the label. The verifier $V_a$ compiles both methods, runs them on $T$, and confirms $\lambda$: the outputs agree on every input for \textsc{eq}, and differ on at least one for \textsc{neq}.

\vspace{-0.06in}
\paragraph{Interaction.}
The relation is type compatibility as in~\cref{sec:verus}, here equality of a pair's shared signature: $\tau_{\mathrm{out}}(e_i) = \tau_{\mathrm{in}}(e_j)$.

\vspace{-0.06in}
\paragraph{Composition.}
For $G=(e_1,\ldots,e_N)$, we apply the same renaming, chaining, and driver
construction as in~\cref{sec:verus} independently to each side of the pairs:
the first methods form one composite, and the second methods form the other.
The task $x_G$ is this pair of programs, and $V_G$ compiles and executes
both against the merged suite.
$\mathrm{Ref}(G)=\textsc{eq}$ when every atom is \textsc{eq}, and
\textsc{neq} when exactly one \textsc{neq} atom is injected at a random position.

\vspace{-0.06in}
\paragraph{Admission.}
Both checks reduce to execution: (1) every atom compiles and its own verifier confirms its label; (2) the composite compiles end-to-end, and executing the merged suite returns $\mathrm{Ref}(G)$, so a positive composition agrees on every test and a negative one diverges on at least one. Groups failing either check are discarded.

\vspace{-0.06in}
\paragraph{Construction.}

Atoms are drawn from FEMPD~\citep{higo2024fempd}, which provides 2,195 Java method pairs judged by three reviewers (1,342 equivalent, 852 not) together with EvoSuite~\citep{fraser2011evosuite} test suites. We re-verify every pair by execution under two independent test-generation methods and keep only pairs both confirm
(\cref{app:clone-construction}). Groups are then grown by the procedure of~\cref{sec:graph} and admitted by execution, at $N \in \{1,2,3,10\}$.
\section{Evaluating frontier models using generated data}
\label{sec:experiments}
 
\vspace{-0.06in}
\subsection{Setup}
To evaluate the effectiveness of \sys, we use it to generate data for the four coding domains above, and measure how challenging the generated tasks are for frontier models as the composition level $N$ increases.

We use our generated data to evaluate Kimi-K3, GPT-5.6-Sol, and Qwen3.5-9B, covering two frontier models and one small open-weight model. Multi-bug repair and proof completion are agentic, while complexity analysis and clone equivalence are single model calls without tool access. Every model uses the same tasks and the same prompt within a domain, and a task is solved when it passes the same verifier used at
admission; runs that exhaust their budget or return no parseable answer count as failures. \Cref{app:setup} summarizes the details.


\begin{table}[t]
\centering
\newsavebox{\evalResultsBox}
\newsavebox{\rftResultsBox}
\newlength{\rftBodyRowPad}
\newcommand{\evalResultsTabular}{%
\begin{tabular}{@{}lcccc@{}}
\toprule
Model & $N{=}1$ & $N{=}2$ & $N{=}3$ & $N{=}10$ \\
\midrule
\multicolumn{5}{@{}l}{\emph{Multi-bug repair} (200/200/168 tasks)} \\
Qwen3.5-9B  & 43.5 & 17.5 & 6.5 & -- \\
Kimi-K3     & 82.0 & 61.5 & 43.5 & -- \\
GPT-5.6-Sol & 100.0 & 86.5 & 78.6 & -- \\
\addlinespace
\multicolumn{5}{@{}l}{\emph{Proof completion} (100/100/100/100 tasks)} \\
Qwen3.5-9B  & 12.0 & 0.0 & 0.0 & 0.0 \\
Kimi-K3     & 87.9 & 69.6 & 57.9 & 0.0 \\
GPT-5.6-Sol & 94.0 & 69.0 & 35.0 & 0.0 \\
\addlinespace
\multicolumn{5}{@{}l}{\emph{Complexity analysis} (80/80/80/80 tasks)} \\
Qwen3.5-9B  & 53.8 & 3.8 & 1.3 & 1.3 \\
Kimi-K3     & 63.8 & 60.0 & 48.8 & 38.8 \\
GPT-5.6-Sol & 68.8 & 65.0 & 57.5 & 35.0 \\
\addlinespace
\multicolumn{5}{@{}l}{\emph{Clone equivalence} (86/82/84/161--166 tasks)} \\
Qwen3.5-9B  & 79.1 & 57.3 & 50.0 & 10.8 \\
Kimi-K3     & 95.3 & 87.8 & 90.5 & 73.3 \\
GPT-5.6-Sol & 98.8 & 90.2 & 89.3 & 72.3 \\
\bottomrule
\end{tabular}
}
\newcommand{\rftResultsTabular}{%
\begin{tabular}{@{}lrrr@{}}
\toprule
Checkpoint & L1 & L2 & L3 \\
\midrule
\multicolumn{4}{@{}l}{\emph{Multi-bug repair}} \\
\noalign{\vskip\rftBodyRowPad}
Base & 44.6 & 18.3 & 7.4 \\
\noalign{\vskip\rftBodyRowPad}
RFT-L1 & 63.4 & 42.7 & 8.7 \\
\noalign{\vskip\rftBodyRowPad}
RFT-L1+L2 & 65.0 & 46.3 & 11.5 \\
\noalign{\vskip\rftBodyRowPad}
RFT-L1+L2+L3 & 66.9 & 52.4 & 12.5 \\
\noalign{\vskip\rftBodyRowPad}
\addlinespace
\multicolumn{4}{@{}l}{\emph{Clone equivalence}} \\
\noalign{\vskip\rftBodyRowPad}
Base & 76.6 & 68.0 & 38.9 \\
\noalign{\vskip\rftBodyRowPad}
RFT-L1 & 93.1 & 86.9 & 53.8 \\
\noalign{\vskip\rftBodyRowPad}
RFT-L1+L2 & 96.2 & 88.9 & 54.6 \\
\noalign{\vskip\rftBodyRowPad}
RFT-L1+L2+L3 & 96.4 & 89.5 & 56.3 \\
\bottomrule
\end{tabular}
}
\small
\setlength{\tabcolsep}{3pt}
\setlength{\rftBodyRowPad}{0pt}
\sbox{\evalResultsBox}{\evalResultsTabular}
\sbox{\rftResultsBox}{\rftResultsTabular}
\setlength{\rftBodyRowPad}{\dimexpr
  \ht\evalResultsBox+\dp\evalResultsBox-
  \ht\rftResultsBox-\dp\rftResultsBox\relax}
\divide\rftBodyRowPad by 9
\sbox{\rftResultsBox}{\rftResultsTabular}
\begin{minipage}[t]{0.53\textwidth}
\vspace{0pt}
\centering
\small
\setlength{\tabcolsep}{3pt}
\usebox{\evalResultsBox}
\caption{Success rate (\%) by composition level. Multi-bug repair runs to $N{=}3$ (\cref{app:bug-construction}); the other domains extend to $N{=}10$. Task counts refer to the scheduled task banks. Proof-completion rates for Kimi-K3 (all levels) and GPT-5.6-Sol (N=10) are over tasks retained after excluding runner failures.}
\label{tab:results}
\end{minipage}%
\hfill
\begin{minipage}[t]{0.44\textwidth}
\vspace{0pt}
\centering
\small
\setlength{\tabcolsep}{3pt}
\usebox{\rftResultsBox}
\caption{Per-attempt success (\%) of Qwen3.5-9B before and after RFT, over eight attempts per task.}
\label{tab:rft-success}
\end{minipage}
\vspace{-0.1in}
\end{table}

\vspace{-0.06in}
\subsection{Multi-bug repair}
\label{sec:exp-repair}
 
\vspace{-0.06in}
\paragraph{Results.}
Solve rates fall with $N$ for every model (\cref{tab:results}): from $N{=}1$ to $N{=}3$, Qwen3.5-9B falls from $43.5\%$ to $6.5\%$, Kimi-K3 from $82.0\%$ to $43.5\%$, and GPT-5.6-Sol from $100.0\%$ to $78.6\%$. The three
models rank in the same order at every level, with large gaps between them, so the benchmark is not saturated even for the strongest model. GPT-5.6-Sol solves every single-bug task, so the source tasks alone cannot distinguish it from a perfect solver; co-locating the same bugs lowers it to $78.6\%$. Rates are higher than on the SWE-rebench leaderboard because we grade on $\mathcal{F}$ alone (\cref{sec:repair}, \cref{app:bug-construction}). We stop at $N{=}3$ because too few groups exist at higher levels in the repositories we mined (\cref{app:bug-construction}).
 
\vspace{-0.04in}
\paragraph{Why accuracy declines.}
We attribute the decline to three compounding factors.
\emph{(i) Conjunctive success criterion.} A level-$N$ task is solved only if all $N$ repairs succeed, so independent per-bug success probability $p$ yields a joint rate of roughly $p^{N}$. Using the $N{=}1$ rate as $p$, this baseline predicts $19/8\%$ for Qwen3.5-9B at $N{=}2/3$, close to the observed $17.5/6.5\%$. Kimi-K3 falls below its baseline at both levels, and GPT-5.6-Sol, whose baseline is $100\%$ throughout, falls furthest below. Independent failure therefore accounts for the weak model's decline but not the strong models.
\emph{(ii) Localization load.} The agent must recover $N$ root causes from a single merged failing-test signal $\mathcal{F}$, without per-issue descriptions and without being told $N$.
\emph{(iii) Interference.} Because coupled bugs share a file by construction, an edit for one defect can perturb the code region of another, so repairs cannot always be composed independently. This pushes difficulty beyond the $p^{N}$ baseline, most sharply for models that fix bugs in isolation but do not reconcile their interactions.
\vspace{-0.04in}
\paragraph{Failures of strong models are composition-induced.}
To separate difficulty due to co-location from difficulty due to individual bugs, we use the fact that every bug in a task lies at the same base commit.
For a level-$N$ task whose constituents each have an isolated ($N{=}1$) evaluation, we classify a failure as an \emph{interference error} if every constituent is solved in isolation but the joint task fails. On the subset with full $N{=}1$ coverage, the interference share of failures rises with model strength: $8\%$ for Qwen3.5-9B, $39\%$ for Kimi-K3, and $100\%$ for GPT-5.6-Sol. Co-location, rather than per-bug difficulty, is therefore the dominant failure mode for exactly the models that fix the bugs individually.

\vspace{-0.06in}
\subsection{Proof completion}
\label{sec:exp-verus}

\vspace{-0.04in}
\paragraph{Results.}
Proof completion becomes harder as more functions are connected
(\cref{tab:results}): from $N{=}1$ to $N{=}3$, Qwen3.5-9B falls from
$12.0\%$ to $0.0\%$, Kimi-K3 from $87.9\%$ to $57.9\%$, and GPT-5.6-Sol from
$94.0\%$ to $35.0\%$. No model solves any task at $N{=}10$. 

\vspace{-0.04in}
\paragraph{Failure analysis.}
Proving a chain requires more than proving each function separately:
each function's contract must also justify subsequent calls and the assertions in the driver (\cref{sec:verus}). For comparison, an illustrative baseline that treats
function-level successes as independent predicts $88\%$ and $83\%$
for GPT at $N{=}2,3$, well above the observed rates.
Among GPT verification
failures, $55\%$ at $N{=}2$ and $70\%$ at $N{=}3$ include errors both in
the driver and elsewhere in the program. All of these verification
failures reach the tenth repair round.
At $N{=}10$, 19 of the GPT tasks produce candidates that Verus accepts, but all 19 omit required specification blocks and therefore fail the final check: they verify a simplified fragment rather than the chained program, so no task counts as solved.

\vspace{-0.06in}

\subsection{Complexity analysis}
\label{sec:exp-complexity}

\vspace{-0.04in}
\paragraph{Results.}
Accuracy decreases as composition depth $N$ increases across all models (\cref{tab:results}), though the degradation trajectories diverge significantly. 
From $N{=}1$ to $N{=}10$, Kimi-K3 falls from $63.8\%$ to
$38.8\%$ and GPT-5.6-Sol from $68.8\%$ to $35.0\%$, a gradual decline. Qwen3.5-9B instead collapses after a single composition step, from $53.8\%$ to $3.8\%$ at $N{=}2$, and stays near zero thereafter.

\vspace{-0.04in}

\paragraph{Why accuracy declines.}
Evaluating end-to-end complexity requires tracking not only each stage's cost but the size of the input it receives, which depends on the output growth of the stages before it: a stage that is $\mathcal{O}(n\log n)$ becomes $\mathcal{O}(n^2\log n)$ when its predecessor emits $\Theta(n^2)$ elements from an input of size $n$. A single mistake about a predecessor's growth invalidates the input size of every stage after it, so errors compound along the chain.

\vspace{-0.04in}

\paragraph{Composition exhausts the small model's budget.}
Qwen3.5-9B's failures at $N \geq 2$ are almost entirely timeouts: under the same budget that suffices at $N{=}1$, and that the frontier models stay within at every level, the model no longer produces a complexity class at all. Composition therefore does not merely make the answer harder to get right for a small model;
it makes the problem too long to finish.

\vspace{-0.06in}
\subsection{Clone equivalence}
\label{sec:clone-results}

\vspace{-0.04in}
\paragraph{Results.}
Both frontier models start near ceiling and lose roughly a quarter of their accuracy by $N{=}10$ (\cref{tab:results}): Kimi-K3 falls from $95.3\%$ to $73.3\%$ and GPT-5.6-Sol from $98.8\%$ to $72.3\%$. The two are close at every level, so the effect does not depend on a particular model. Qwen3.5-9B declines
far faster, from $79.1\%$ to $10.8\%$, ending well below the $50\%$ a coin flip would give on a balanced set.

\paragraph{The trend is visible only over the full range.}
Adjacent levels are not distinguishable: $N{=}2$ and $N{=}3$ differ by less than the per-level noise for both frontier models, at every dataset scale we tried. We therefore run to $N{=}10$, an order of magnitude beyond the single-pair baseline, which is far enough for the effect to be unambiguous and still within what the 127 type-composable negative atoms support without shrinking per-level samples.

\vspace{-0.04in}
\paragraph{Why accuracy declines.}
The task has a binary answer, so a model that ignores composition still scores $50\%$, and the near-ceiling accuracy at low $N$ is expected: a single injected non-equivalent pair is usually visible locally. The two directions are not equally hard, though. Negative-class accuracy stays near ceiling at every level
for both frontier models, while positive-class accuracy falls steeply, from above $97\%$ at $N{=}1$ to $46\%$ and $51\%$ at $N{=}10$. Spotting one diverging pair is a local check, whereas certifying equivalence requires all $N$ pairs to be equivalent and no stage to mask an upstream difference. As depth grows, models increasingly answer \textsc{neq} even when no divergence exists, which is what drives the decline. \Cref{app:clone-class} gives the per-class accuracies.

\section{Fine-Tuning Using Generated Data}
\label{sec:rft}

We apply agentic RFT to Qwen3.5-9B using \sys-generated tasks; the experimental setup is detailed in ~\cref{app:rft-hyperparameters}.
Since our goal is to show that the generated tasks can train coding agents, both domains run in the same agent scaffold, including clone equivalence, which~\cref{tab:results}
evaluates with a single model call, and results are reported on a held-out split, so Base rates differ from~\cref{tab:results}.


The combined tasks generated by \sys are useful not only for evaluation but also as training data for coding agents. As shown in ~\cref{tab:rft-success}, RFT on the full L1+L2+L3 dataset improves Qwen3.5-9B over Base at every composition level in both domains. On L3 tasks, success rises from $7.4\%$ to $12.5\%$ for Multi-bug repair and from $38.9\%$ to $56.3\%$ for Clone equivalence. These gains show that the generated datasets support learning to solve composed tasks through agentic RFT.

Higher-level combined tasks also provide value beyond training on atomic tasks alone. Compared with \mbox{RFT-L1}, the full curriculum yields further gains of $3.5/9.7/3.8$ percentage points on Multi-bug repair and $3.3/2.6/2.5$ points on Clone equivalence at L1/L2/L3. The gains extend across all three levels: training with combined tasks improves performance on complex compositions while retaining and further improving performance on atomic tasks. Together, these results support using composition to expand a pool of atomic problems into effective training data for coding agents.

\section{Related Work}



\label{sec:related}

\paragraph{Coding benchmarks and how they are built.}
Prompt-and-test benchmarks~\citep{chen2021humaneval, austin2021mbpp} gave way to
repository-level tasks mined with the tests that certify their
fixes~\citep{jimenez2024swebench}, and to pipelines that automate collection so
the pool can be refreshed and contamination
tracked~\citep{badertdinov2025swerebench, badertdinov2026swerebenchv2}. A
parallel line synthesizes tasks instead, injecting bugs or building executable
environments around mined instances~\citep{yang2025swesmith, pan2024swegym,
jain2025r2egym}; these rely on a language model to propose and often to validate
each task, which bounds difficulty by what the generator produces. Both lines
yield pools of atomic problems whose difficulty is fixed at collection time. We
treat such a pool as atoms: our generator is a deterministic program over
verified material, and difficulty is set by the composition level rather than by
what was mined or generated. The benchmarks supplying our atoms are of this
kind~\citep{baik2025codecomplex, higo2024fempd, yang2025autoverus}.

\paragraph{Composition.}
Models solve sub-problems more reliably than their composition: they answer
sub-questions correctly yet fail to combine them~\citep{press2023measuring},
degrade with the number of composed operations~\citep{dziri2023faith}, and lose
accuracy when a problem is lengthened~\citep{mirzadeh2024gsmsymbolic}. These
studies diagnose composition as a weakness, mostly on symbolic tasks with short
answers. Two recent efforts compose coding problems.
\citet{fang2026playingpsychic} compose existing problems to obtain
contamination-resistant evaluation sets for an analysis of reasoning traces on
one family of tasks. \citet{jin2026chainswe} chain issues so that an agent fixes
bugs one after another, and attribute later failures to earlier repairs. Our
tasks are simultaneous rather than sequential: all $N$ atomic problems are
unsolved in one program, with no ordering, no per-problem prompt, and no
disclosure of $N$. Interference errors are the simultaneous analogue of their
chain errors, and we study composition across four domains under one formulation
in which a domain is added by defining a single relation.

\section{Conclusion}
We introduced \sys, a model-free framework that generates coding benchmarks by composing verified atomic problems into tasks of arbitrary level of difficulty. Using \sys-generated data, we show that frontier models degrade substantially with depth, and this degradation is accompanied by substantial composition-induced interference rather than being explained solely by failures on the individual constituent problems. \sys-generated data also serves as effective training signal: agentic rejection-sampling fine-tuning improves Qwen3.5-9B by up to 34.1 percentage points over the base model. These results establish \sys as a scalable mechanism for building benchmarks that probe compositional reasoning beyond what atomically saturated benchmarks can measure.


\bibliographystyle{plainnat}
\bibliography{references_codewalk}

@article{yang2025autoverus,
  title         = {{AutoVerus}: Automated Proof Generation for {Rust} Code},
  author        = {Yang, Chenyuan and Li, Xuheng and Misu, Md Rakib Hossain and Yao, Jianan and Cui, Weidong and Gong, Yeyun and Hawblitzel, Chris and Lahiri, Shuvendu K. and Lorch, Jacob R. and Lu, Shuai and Yang, Fan and Zhou, Ziqiao and Lu, Shan},
  journal       = {Proceedings of the ACM on Programming Languages},
  volume        = {9},
  number        = {OOPSLA2},
  pages         = {3454--3482},
  year          = {2025},
  doi           = {10.1145/3763174},
  eprint        = {2409.13082},
  archivePrefix = {arXiv},
  url           = {https://arxiv.org/abs/2409.13082}
}

@article{jimenez2024swebench,
  title         = {{SWE-bench}: Can Language Models Resolve Real-World {GitHub} Issues?},
  author        = {Jimenez, Carlos E. and Yang, John and Wettig, Alexander and Yao, Shunyu and Pei, Kexin and Press, Ofir and Narasimhan, Karthik},
  journal       = {arXiv preprint arXiv:2310.06770},
  year          = {2023},
  eprint        = {2310.06770},
  archivePrefix = {arXiv},
  url           = {https://arxiv.org/abs/2310.06770},
  note          = {Published at ICLR 2024}
}

@article{badertdinov2025swerebench,
  title         = {{SWE-rebench}: An Automated Pipeline for Task Collection and Decontaminated Evaluation of Software Engineering Agents},
  author        = {Badertdinov, Ibragim and Golubev, Alexander and Nekrashevich, Maksim and Shevtsov, Anton and Karasik, Simon and Andriushchenko, Andrei and Trofimova, Maria and Litvintseva, Daria and Yangel, Boris},
  journal       = {arXiv preprint arXiv:2505.20411},
  year          = {2025},
  eprint        = {2505.20411},
  archivePrefix = {arXiv},
  url           = {https://arxiv.org/abs/2505.20411}
}

@article{badertdinov2026swerebenchv2,
  title         = {{SWE-rebench V2}: Language-Agnostic {SWE} Task Collection at Scale},
  author        = {Badertdinov, Ibragim and Nekrashevich, Maksim and Shevtsov, Anton and Golubev, Alexander},
  journal       = {arXiv preprint arXiv:2602.23866},
  year          = {2026},
  eprint        = {2602.23866},
  archivePrefix = {arXiv},
  url           = {https://arxiv.org/abs/2602.23866}
}

@article{yang2025swesmith,
  title         = {{SWE-smith}: Scaling Data for Software Engineering Agents},
  author        = {Yang, John and Lieret, Kilian and Jimenez, Carlos E. and Wettig, Alexander and Khandpur, Kabir and Zhang, Yanzhe and Hui, Binyuan and Press, Ofir and Schmidt, Ludwig and Yang, Diyi},
  journal       = {arXiv preprint arXiv:2504.21798},
  year          = {2025},
  eprint        = {2504.21798},
  archivePrefix = {arXiv},
  url           = {https://arxiv.org/abs/2504.21798}
}

@article{pan2024swegym,
  title         = {Training Software Engineering Agents and Verifiers with {SWE-Gym}},
  author        = {Pan, Jiayi and Wang, Xingyao and Neubig, Graham and Jaitly, Navdeep and Ji, Heng and Suhr, Alane and Zhang, Yizhe},
  journal       = {arXiv preprint arXiv:2412.21139},
  year          = {2024},
  eprint        = {2412.21139},
  archivePrefix = {arXiv},
  url           = {https://arxiv.org/abs/2412.21139}
}

@article{jain2025r2egym,
  title         = {{R2E-Gym}: Procedural Environments and Hybrid Verifiers for Scaling Open-Weights {SWE} Agents},
  author        = {Jain, Naman and Singh, Jaskirat and Shetty, Manish and Zheng, Liang and Sen, Koushik and Stoica, Ion},
  journal       = {arXiv preprint arXiv:2504.07164},
  year          = {2025},
  eprint        = {2504.07164},
  archivePrefix = {arXiv},
  url           = {https://arxiv.org/abs/2504.07164}
}

@article{yang2024sweagent,
  title         = {{SWE-agent}: Agent-Computer Interfaces Enable Automated Software Engineering},
  author        = {Yang, John and Jimenez, Carlos E. and Wettig, Alexander and Lieret, Kilian and Yao, Shunyu and Narasimhan, Karthik and Press, Ofir},
  journal       = {arXiv preprint arXiv:2405.15793},
  year          = {2024},
  eprint        = {2405.15793},
  archivePrefix = {arXiv},
  url           = {https://arxiv.org/abs/2405.15793}
}

@article{jin2026chainswe,
  title         = {{ChainSWE}: Benchmarking Coding Agents on Multi-Bug Software Maintenance},
  author        = {Jin, Qirui and Tung, Lingching and Li, Kenan and Shi, Qiyang and She, Yushi and Jia, Huanzhong and Zhao, Harrison and Xia, Kejing and Du, Zhenbang and Zhang, Yikai and Pei, Jiaxin and Zhang, Zhenyu and Qi, Zhen and Duan, Yuyan and Lee, Wenke and Jin, Zijian},
  journal       = {arXiv preprint arXiv:2607.02606},
  year          = {2026},
  eprint        = {2607.02606},
  archivePrefix = {arXiv},
  url           = {https://arxiv.org/abs/2607.02606}
}

@inproceedings{fraser2011evosuite,
  author    = {Fraser, Gordon and Arcuri, Andrea},
  title     = {{EvoSuite}: Automatic Test Suite Generation for Object-Oriented Software},
  booktitle = {Proceedings of the 19th ACM SIGSOFT Symposium and the 13th European Conference on Foundations of Software Engineering (ESEC/FSE '11)},
  pages     = {416--419},
  year      = {2011},
  doi       = {10.1145/2025113.2025179}
}

@inproceedings{press2023measuring,
  title         = {Measuring and Narrowing the Compositionality Gap in Language Models},
  author        = {Press, Ofir and Zhang, Muru and Min, Sewon and Schmidt, Ludwig and Smith, Noah A. and Lewis, Mike},
  booktitle     = {Findings of the Association for Computational Linguistics: EMNLP 2023},
  year          = {2023},
  eprint        = {2210.03350},
  archivePrefix = {arXiv},
  url           = {https://arxiv.org/abs/2210.03350}
}

@inproceedings{dziri2023faith,
  title         = {Faith and Fate: Limits of Transformers on Compositionality},
  author        = {Dziri, Nouha and Lu, Ximing and Sclar, Melanie and Li, Xiang Lorraine and Jiang, Liwei and Lin, Bill Yuchen and West, Peter and Bhagavatula, Chandra and Le Bras, Ronan and Hwang, Jena D. and Sanyal, Soumya and Welleck, Sean and Ren, Xiang and Ettinger, Allyson and Harchaoui, Zaid and Choi, Yejin},
  booktitle     = {Advances in Neural Information Processing Systems},
  volume        = {36},
  year          = {2023},
  eprint        = {2305.18654},
  archivePrefix = {arXiv},
  url           = {https://arxiv.org/abs/2305.18654}
}

@article{mirzadeh2024gsmsymbolic,
  title         = {{GSM-Symbolic}: Understanding the Limitations of Mathematical Reasoning in Large Language Models},
  author        = {Mirzadeh, Iman and Alizadeh, Keivan and Shahrokhi, Hooman and Tuzel, Oncel and Bengio, Samy and Farajtabar, Mehrdad},
  journal       = {arXiv preprint arXiv:2410.05229},
  year          = {2024},
  eprint        = {2410.05229},
  archivePrefix = {arXiv},
  url           = {https://arxiv.org/abs/2410.05229}
}

@article{fang2026playingpsychic,
  title         = {Playing Psychic: Using Thought Trees to Predict Reasoning Models Accuracy on Coding Tasks},
  author        = {Fang, Jiaxin and He, Runyuan and Bhatia, Sahil and Gajare, Neel and Cheung, Alvin},
  journal       = {arXiv preprint arXiv:2604.16931},
  year          = {2026},
  eprint        = {2604.16931},
  archivePrefix = {arXiv},
  url           = {https://arxiv.org/abs/2604.16931}
}

@article{chen2021humaneval,
  title         = {Evaluating Large Language Models Trained on Code},
  author        = {Chen, Mark and Tworek, Jerry and Jun, Heewoo and Yuan, Qiming and Pinto, Henrique Ponde de Oliveira and Kaplan, Jared and Edwards, Harri and Burda, Yuri and Joseph, Nicholas and Brockman, Greg and others},
  journal       = {arXiv preprint arXiv:2107.03374},
  year          = {2021},
  eprint        = {2107.03374},
  archivePrefix = {arXiv},
  url           = {https://arxiv.org/abs/2107.03374}
}

@article{austin2021mbpp,
  title         = {Program Synthesis with Large Language Models},
  author        = {Austin, Jacob and Odena, Augustus and Nye, Maxwell and Bosma, Maarten and Michalewski, Henryk and Dohan, David and Jiang, Ellen and Cai, Carrie and Terry, Michael and Le, Quoc and Sutton, Charles},
  journal       = {arXiv preprint arXiv:2108.07732},
  year          = {2021},
  eprint        = {2108.07732},
  archivePrefix = {arXiv},
  url           = {https://arxiv.org/abs/2108.07732}
}

@article{dai2024mhpp,
  title         = {{MHPP}: Exploring the Capabilities and Limitations of Language Models Beyond Basic Code Generation},
  author        = {Dai, Jianbo and Lu, Jianqiao and Feng, Yunlong and Zeng, Guangtao and Ruan, Rongju and Cheng, Ming and Huang, Dong and Tan, Haochen and Guo, Zhijiang},
  journal       = {arXiv preprint arXiv:2405.11430},
  year          = {2024},
  eprint        = {2405.11430},
  archivePrefix = {arXiv},
  url           = {https://arxiv.org/abs/2405.11430}
}

@article{jain2024livecodebench,
  title         = {{LiveCodeBench}: Holistic and Contamination Free Evaluation of Large Language Models for Code},
  author        = {Jain, Naman and Han, King and Gu, Alex and Li, Wen-Ding and Yan, Fanjia and Zhang, Tianjun and Wang, Sida and Solar-Lezama, Armando and Sen, Koushik and Stoica, Ion},
  journal       = {arXiv preprint arXiv:2403.07974},
  year          = {2024},
  eprint        = {2403.07974},
  archivePrefix = {arXiv},
  url           = {https://arxiv.org/abs/2403.07974}
}

@inproceedings{baik2025codecomplex,
  title         = {{CodeComplex}: Dataset for Worst-Case Time Complexity Prediction},
  author        = {Baik, SeungYeop and Hahn, Joonghyuk and Kim, Jungin and Aditi and Jeon, Mingi and Han, Yo-Sub and Ko, Sang-Ki},
  booktitle     = {Findings of the Association for Computational Linguistics: EMNLP 2025},
  pages         = {19616--19638},
  year          = {2025},
  eprint        = {2401.08719},
  archivePrefix = {arXiv},
  url           = {https://arxiv.org/abs/2401.08719}
}

@article{higo2024fempd,
  title   = {Dataset of Functionally Equivalent {Java} Methods and Its Application to Evaluating Clone Detection Tools},
  author  = {Higo, Yoshiki},
  journal = {IEICE Transactions on Information and Systems},
  volume  = {E107.D},
  number  = {6},
  pages   = {751--760},
  year    = {2024},
  doi     = {10.1587/transinf.2023EDP7268}
}

@inproceedings{svajlenko2014bigclonebench,
  title     = {Towards a Big Data Curated Benchmark of Inter-Project Code Clones},
  author    = {Svajlenko, Jeffrey and Islam, Judith F. and Keivanloo, Iman and Roy, Chanchal K. and Mia, Mohammad Mamun},
  booktitle = {IEEE International Conference on Software Maintenance and Evolution (ICSME)},
  pages     = {476--480},
  year      = {2014},
  doi       = {10.1109/ICSME.2014.77}
}

\appendix
\newpage
\appendix
\crefalias{section}{appendix}
\crefalias{subsection}{appendix}
\crefalias{subsubsection}{appendix}

\section{Dataset examples}
\label{app:examples}

Each subsection gives one composed task, its reference answer, and the prompt
the model receives.

\subsection{Multi-bug repair}

\begin{figure}[h]\small
  \textbf{Bug A (\texttt{tabulate-96}) --- gold patch}
  \begin{lstlisting}[language=diff,basicstyle=\ttfamily\scriptsize]
--- a/tabulate.py
+++ b/tabulate.py
@@ def _normalize_tabular_data(tabular_data, headers, showindex="default"):
             keys = list(tabular_data)
-            if tabular_data.index.name is not None:
+            if showindex in ["default", "always", True] and tabular_data.index.name is not None:
                 if isinstance(tabular_data.index.name, list):
  \end{lstlisting}
  \textbf{Bug A --- test patch}
  \begin{lstlisting}[language=diff,basicstyle=\ttfamily\scriptsize]
--- a/test/test_output.py
+++ b/test/test_output.py
@@ def test_pandas_without_index():
-            [["one", 1], ["two", None]], columns=["string", "number"], index=["a", "b"]
+            [["one", 1], ["two", None]], columns=["string", "number"],
+            index=pandas.Index(["a", "b"], name="index")
  \end{lstlisting}
  \textbf{Bug B (\texttt{tabulate-139}) --- gold patch}
  \begin{lstlisting}[language=diff,basicstyle=\ttfamily\scriptsize]
--- a/tabulate.py
+++ b/tabulate.py
@@ def _normalize_tabular_data(tabular_data, headers, showindex="default"):
         rows = rows[1:]
+    elif headers == "firstrow":
+        headers = []

     headers = list(map(_text_type, headers))
  \end{lstlisting}
  \textbf{Bug B --- test patch}
  \begin{lstlisting}[language=diff,basicstyle=\ttfamily\scriptsize]
--- a/test/test_output.py
+++ b/test/test_output.py
@@ def test_empty_data_without_headers():
+def test_empty_data_with_headers():
+    "Output: table with empty data and headers as firstrow"
+    expected = ""
+    result = tabulate([], headers="firstrow")
+    assert_equal(expected, result)
  \end{lstlisting}
\caption{A level-2 co-located task. Two fixes to
\texttt{astanin/python-tabulate}, both editing
\texttt{\_normalize\_tabular\_data} in \texttt{tabulate.py}, are present together
at the earlier commit $c_\star$. The agent receives the repository with both test
patches applied, so both tests fail, and must edit the source so both pass.
Admission check~(1): each test fails at $c_\star$; check~(2): applying both gold
patches makes both pass.}
\label{fig:example-repair}
\end{figure}

\noindent\textbf{Prompt:}
\begin{framed}
\begin{verbatim}
A Python repo is at /testbed (conda env `testbed` active). There are
MULTIPLE bugs in the library source. These tests FAIL:
  {test1}
  {test2}
  ...

Fix the source so ALL of them pass. Note there may be more than one
independent bug to fix. Use the bash tool, one command per turn (fresh
shell, so prefix `cd /testbed && ...`). Run tests with
`cd /testbed && python -m pytest <node id> -q`. Do NOT edit tests.
When all pass, run `echo COMPLETE_TASK_AND_SUBMIT`.
\end{verbatim}
\end{framed}

\subsection{Complexity analysis}

\noindent\textbf{Composed level-2 program:}
\begin{framed}
\begin{verbatim}
def solve_1(a: int, b: int) -> int:
    if a > b:
        a, b = b, a
    ba = bin(a)[2:]
    bb = bin(b)[2:]
    r = ''
    if len(ba) != len(bb):
        int('1' * len(bb), 2)
    else:
        for ca, cb in zip(ba, bb):
            if ca == cb:
                r += '0'
            else:
                r += '1'
                break
    r += '1' * (len(bb) - len(r))
    return int(r, 2)


def solve_2(x: int, n: int, m: int, grid: list) -> bool:
    global ans
    dp = {}
    for i in range(n):
        temp = 0
        for j in range(m):
            if grid[i][j] >= x:
                temp = temp | (1 << j)
        dp[temp] = i
    for aa, bb in dp.items():
        for cc, dd in dp.items():
            if aa | cc == 2 ** m - 1:
                ans = (bb + 1, dd + 1)
                return True
    return False


a, b = map(int, input().split())
ans = (-1, -1)
n, m = map(int, input().split())
grid = []
for i in range(n):
    grid.append(list(map(int, input().split())))
x = solve_1(a, b)
result = solve_2(x, n, m, grid)
print(result)
print(*ans)
\end{verbatim}
\end{framed}

\noindent\textbf{Reference:} \texttt{quadratic}

\noindent\textbf{Prompt:}
\begin{framed}
\begin{verbatim}
Determine the time complexity of this competitive programming code.

Treat "n" as the primary input size (the value read as n, or the length
of the main input array/string).

Judge the complexity of the whole program, not just one function.

Choose exactly one label:

constant
logn
linear
nlogn
quadratic
cubic
np

Return ONLY the label.
Do not explain your answer.
\end{verbatim}
\end{framed}

\begingroup
\raggedbottom
\subsection{Clone equivalence}
\label{app:clone-example}
This illustrative level-2 composition first pads or truncates a string to a
specified width, then checks whether the resulting string contains a query.
The first pair returns a \texttt{String}, which is passed directly to the
second pair's first \texttt{String} argument; the query becomes an additional
driver input. On each side, the driver \texttt{outer} chains that side's
methods, as in~\cref{sec:clone}; method names are kept because they do not
collide within a snippet.

\noindent\textbf{System prompt} (identical across every level and model; the user
prompt is the two composed programs):
\begin{framed}

\begin{verbatim}
You are a program equivalence judge. You will be shown two Java code
snippets, each possibly composed of several helper classes and an outer
method that calls them in sequence. Decide whether the two snippets are
semantically equivalent (compute the same results / have the same
observable behavior for the same inputs to the outer method), regardless
of variable/class names or formatting. Respond with exactly one line:
VERDICT: 1 (equivalent) or VERDICT: 0 (not equivalent).
\end{verbatim}
\end{framed}

\noindent\begin{minipage}[t]{0.49\linewidth}
\textbf{Snippet A:}
\begin{lstlisting}[language=Java,basicstyle=\ttfamily\footnotesize,columns=fullflexible,keepspaces=true,upquote=true,showstringspaces=false,frame=single,framerule=0.3pt,breaklines=true,aboveskip=4pt,belowskip=0pt, literate={"}{{\char34}}1]
class SnippetA {
  static String fixedStringLength(
      String s, int length) {
    if (s.length() < length) {
      StringBuffer str =
          new StringBuffer(s);
      for (int i = 0;
           i < length - s.length(); i++) {
        str.append(" ");
      }
      s = str.toString();
    } else if (s.length() > length) {
      s = s.substring(0, length);
    }
    return s;
  }
  static boolean contains(
      String str, String segment) {
    if ((str == null) || (segment == null))
      return false;
    if (segment.length() == 0) return true;
    return str.indexOf(segment) >= 0;
  }
  static boolean outer(
      String s, int width, String needle) {
    String intermediate =
        fixedStringLength(s, width);
    return contains(intermediate, needle);
  }
}
\end{lstlisting}
\end{minipage}\hfill
\begin{minipage}[t]{0.49\linewidth}
\textbf{Snippet B:}
\begin{lstlisting}[language=Java,basicstyle=\ttfamily\footnotesize,columns=fullflexible,keepspaces=true,upquote=true,showstringspaces=false,frame=single,framerule=0.3pt,breaklines=true,aboveskip=4pt,belowskip=0pt]
class SnippetB {
  static String left(String s, int w) {
    int n = s.length();
    if (w < n) return s.substring(0, w);
    StringBuilder result =
        new StringBuilder(s);
    for (int i = n; i < w; i++) {
      result.append(' ');
    }
    return result.toString();
  }
  static boolean contains(
      String str, String searchStr) {
    if (str == null || searchStr == null) {
      return false;
    }
    return str.indexOf(searchStr) >= 0;
  }
  static boolean outer(
      String s, int width, String needle) {
    String intermediate = left(s, width);
    return contains(intermediate, needle);
  }
}
\end{lstlisting}
\end{minipage}

\paragraph{Reference.}
\textsc{eq}, or \texttt{VERDICT: 1}. The first methods both pad with spaces or
truncate to the requested width; the second methods perform the same
null-safe substring check. Executing the atomic pairs and the full chains
confirms the expected label on the example's tests.
For \mbox{\texttt{outer("ab", 4, " ")}}, both sides return \texttt{true} because padding
introduces spaces. For \mbox{\texttt{outer("abcde", 3, "d")}}, both return \texttt{false}
because truncation removes the queried character. In both cases, the final
answer depends on the first atom's output.
\subsection{Proof completion}
\label{app:verus-example}
This level-2 task chains \texttt{extract\_rear\_chars}, which takes the last byte of each nonempty byte string, with \texttt{contains\_z}, which checks for the ASCII bytes \texttt{Z} or \texttt{z}. For example, the strings \texttt{az} and \texttt{by} produce the intermediate vector \texttt{[z,y]} and the final result \texttt{true}. The generated call uses \texttt{as\_slice()} to pass the first function's output to the second. The example is task \texttt{S2\_024}; the imports, \texttt{main}, and enclosing \texttt{verus!} block are omitted below.

\noindent\begin{minipage}{\linewidth}
\paragraph{Task.}
The component postconditions and proof annotations have been removed. The original nonempty-input precondition remains.
\begin{lstlisting}[style=verus-example,basicstyle=\ttfamily\footnotesize,breaklines=true,morekeywords={spec,requires,forall,while,invariant,decreases,mut}]
fn f1_extract_rear_chars(s: &Vec<Vec<u8>>) -> (result: Vec<u8>)
    requires
        forall|i: int| 0 <= i < s.len() ==> #[trigger] s[i].len() > 0, {
    let mut rear_chars: Vec<u8> = Vec::with_capacity(s.len());
    let mut index = 0;
    while index < s.len() {
        let seq = &s[index];
        rear_chars.push(seq[seq.len() - 1]);
        index += 1;
    }
    rear_chars
}
fn f2_contains_z(text: &[u8]) -> (result: bool) {
    let mut index = 0;
    while index < text.len() {
        if text[index] == 90 || text[index] == 122 {
            return true;
        }
        index += 1;
    }
    false
}
\end{lstlisting}
The fixed predicates \texttt{step1} and \texttt{step2} represent $S_1$ and $S_2$ from \cref{sec:verus}. Here \texttt{v@} denotes the mathematical sequence represented by a vector or slice \texttt{v}. The driver checks both atom specifications on the actual intermediate values.
\begin{lstlisting}[style=verus-example,basicstyle=\ttfamily\footnotesize,breaklines=true,morekeywords={spec,requires,forall,while,invariant,decreases,mut}]
spec fn step1(__hp0: Seq<Vec<u8>>, outp: Seq<u8>) -> bool {
    (__hp0.len() == outp.len())
    && (forall|i: int| 0 <= i < __hp0.len()
        ==> outp[i] == #[trigger] __hp0[i][__hp0[i].len() - 1])
}
spec fn step2(inp: Seq<u8>, outp: bool) -> bool {
    (outp == (exists|i: int| 0 <= i < inp.len()
        && (inp[i] == 90 || inp[i] == 122)))
}
fn chain_driver(s: &Vec<Vec<u8>>) -> (r_final: bool)
    requires
        forall|i: int| 0 <= i < s.len() ==> #[trigger] s[i].len() > 0,
    ensures
        exists|v1: Seq<u8>| #[trigger] step1(s@, v1) && step2(v1, r_final), {
    let x1 = f1_extract_rear_chars(s);
    let x2 = f2_contains_z(x1.as_slice());
    proof {
        assert(step1(s@, x1@));
        assert(step2(x1@, x2));
    }
    x2
}
\end{lstlisting}

\end{minipage}

\noindent\begin{minipage}{\linewidth}
\paragraph{Reference.}
The reference restores each component's original \texttt{ensures} clause and its loop invariants and termination measure. For the first loop, the key invariant states that the output contains exactly the last bytes of the strings processed so far; for the second, it states that no processed byte is \texttt{Z} or \texttt{z}. The following are the restored loop annotations, inserted immediately after the corresponding \texttt{while} conditions.
\begin{lstlisting}[style=verus-example,basicstyle=\ttfamily\footnotesize,breaklines=true,morekeywords={spec,requires,forall,while,invariant,decreases,mut}]
// f1_extract_rear_chars:
invariant
    0 <= index <= s.len(),
    rear_chars.len() == index,
    forall|i: int| 0 <= i < s.len()
        ==> #[trigger] s[i].len() > 0,
    forall|k: int| 0 <= k < index
        ==> rear_chars[k]
            == #[trigger] s[k][s[k].len() - 1],
decreases s@.len() - index,

// f2_contains_z:
invariant
    0 <= index <= text.len(),
    forall|k: int| 0 <= k < index
        ==> (text[k] != 90 && text[k] != 122),
decreases text@.len() - index,
\end{lstlisting}
The restored postconditions express \texttt{step1(s@, result@)} and \texttt{step2(text@, result)}, respectively. Together with these annotations, the reference program verifies, while the uncompleted task does not.

\end{minipage}

\paragraph{Prompt.}
The initial AutoVerus generation prompt includes the following instructions; the full task program is supplied as its code input. This is an excerpt; few-shot examples and subsequent error-guided repair prompts are omitted.
\begin{quote}\small
Your mission is to add loop invariants to the given Rust code, if there are loops in the code, so that Verus can verify the give function behaves exact what is described in the specifications.\\
Respond with Rust code only, do not include any explanation.\\
You should never change or delete existing Rust code.
\end{quote}

\endgroup

\section{Dataset Construction and Experiment Setup}
\label{app:setup}
\subsection{Multi-bug repair}
\label{app:bug-construction}
SWE-rebench~\citep{badertdinov2025swerebench} and
SWE-rebench-V2~\citep{badertdinov2026swerebenchv2} provide the fields an atom requires---base commit, gold patch, test patch, and \texttt{FAIL\_TO\_PASS} tests---and both track contamination against model release dates. Admission is selective at
higher levels: of the candidate groups we attempted, $74\%$ were admitted at $N{=}1$ ($200/270$), $25\%$ at $N{=}2$ ($200/811$), and $11\%$ at $N{=}3$ ($168/1{,}591$). Across the 270 repositories, only 15 groups are available at $N{=}4$ and fewer
than 10 at each of $N{=}5$ through $10$, too few for a reliable measurement. This reflects the number of repositories mined rather than a limit of the generator: the largest level is set by the graph's connectivity, which grows as more repositories are added.

We use a minimal agent loop in the style of SWE agent~\citep{yang2024sweagent}, with a single bash tool invoked through native tool calling and no editor, search, or other interface commands. The prompt gives the repository path and the union $\mathcal{F}$ of failing tests, states that there may be more than one bug, and instructs the agent to fix the source so that all tests in $\mathcal{F}$ pass without editing tests. The prompt does not reveal $N$. Each turn is one model call followed by one command, executed in a fresh shell inside the task container. A run ends when the agent submits, or at 100 turns, a \$1.50 cost limit, or 1500\,s of wall-clock time. The same loop and prompt are used for all models. A task is solved if every test in $\mathcal{F}$ passes after the run. We grade on
$\mathcal{F}$ (the \texttt{FAIL\_TO\_PASS} tests) alone, as our objective is \emph{targeted multi-bug repair}, so $\mathcal{F}$ is the precise oracle for that question. We deliberately
do not additionally gate on the repositories' \texttt{PASS\_TO\_PASS} suites, consequently our solve rate measures bug fixing rather than full-patch acceptability, and is not directly comparable to the ``resolved'' rate of SWE-rebench leaderboard, which additionally requires that no previously passing test regress.

\subsection{Proof completion}
All models use the AutoVerus scaffold~\citep{yang2025autoverus} with its default generation, refinement, and repair phases, up to 10 repair rounds, 16{,}000 tokens per call (4{,}096 for repair calls), and 80 minutes per task. We disable AutoVerus's check that a candidate preserves the input's specifications, since completing a task requires adding postconditions. A task is solved if the final candidate verifies once the task's original stage predicates and driver are restored, and contains no proof-bypass construct (\texttt{assume}, \texttt{admit}, \texttt{external\_body}, \texttt{exec\_allows\_no\_decreases}), so a candidate cannot earn credit by weakening a predicate or rewriting the driver. Timeouts, and candidates that omit a predicate or the driver, count as failures.

\paragraph{Construction details.}
\label{app:verus-construction}
We select functions with named return values and postconditions, then
copy each postcondition into a \texttt{step} predicate representing $S_i$.
Vectors and slices are represented in these predicates by mathematical
sequences of elements using Verus's sequence view.
Runtime connections support direct passing, borrowing, and converting
\texttt{Vec<T>} to \texttt{\&[T]} with \texttt{as\_slice()}.
Other downstream arguments must be scalars and become driver inputs.
The driver's precondition collects the first function's preconditions
and later preconditions independent of the chained input; conditions on
that input must be established by upstream contracts.
Its end-to-end postcondition is
$\exists v_1,\ldots,v_{N-1},\;\bigwedge_{i=1}^{N}S_i(v_{i-1},v_i)$,
where $v_0=x$, $v_N=r$, and additional fixed inputs are omitted from the
notation. The driver also asserts each $S_i$ on the actual intermediate
values. Annotation removal affects component postconditions and proof
annotations, including loop invariants and proof blocks, but preserves
the driver and $S_i$ definitions. Admission checks verification failure
of the resulting whole program, not of each component independently.

We enumerate compatible pairs at $N{=}2$ and sample extensions for
$N{=}3$, excluding duplicate chains. In recorded construction rounds,
Verus accepts $137/160$ ($86\%$), $200/350$ ($57\%$), and $233/770$
($30\%$) candidate references at $N{=}1,2,3$, before annotation removal;
these are reference-verification yields, not final admission rates.
The resulting 300 tasks use 100 distinct source-function pairs across
87 programs. For $N{=}10$, we follow connections from verified shorter
chains, permit each source program at most twice per chain, and apply
both admission checks to each full program to obtain another 100 tasks.

\subsection{Complexity analysis}
\label{app:complex-construction}
CodeComplex~\citep{baik2025codecomplex} contains 4{,}900 Python programs annotated across seven asymptotic time-complexity classes ($\mathcal{O}(1)$, $\mathcal{O}(\log n)$,
$\mathcal{O}(n)$, $\mathcal{O}(n \log n)$, $\mathcal{O}(n^2)$,
$\mathcal{O}(n^3)$, exponential). We group programs by problem identifier and sample one snippet per problem, parsing each into an abstract syntax tree to extract its parameter and return type signatures; keeping only programs with well-defined, composable types yields the pool of 277 atoms. For admission check~(1), $100\%$, $83.4\%$, $56.7\%$, and $3.6\%$ of evaluated candidate
chains were admitted at $N \in \{1,2,3,10\}$, and the 80 tasks per level are drawn from these uniformly at random.
 
We use a simple single-turn direct generation scaffold without tool access, scratchpad loops, or external execution commands. The prompt gives the full source code containing all $N$ constituent functions and instructs the model to select the overall worst-case time complexity bound from a closed set of complexity classes (e.g., $\mathcal{O}(1)$, $\mathcal{O}(n)$, $\mathcal{O}(n \log n)$, $\mathcal{O}(n^2)$). Composition depth $N$ corresponds to the length of the function call chain over which intermediate complexity bounds must be tracked and composed. Each task is evaluated across four depth levels ($N = 1, 2, 3, 10$) with 80 tasks per level (320 tasks per model). A run ends on model completion, or at a 4,096 token limit, or a timeout of 300 s of wall-clock time. The same scaffold and prompt are used for all models. A task is solved if the predicted class matches the ground-truth complexity label. We extract the predicted class via automated string parsing; runs that time out or yield outputs without a parseable complexity class are scored as incorrect.

\subsection{Clone equivalence}
\label{app:clone-construction}
Atoms are drawn from FEMPD~\citep{higo2024fempd}. We re-verify every pair by execution under two independent test-generation methods, type-driven generation and cross-substitution of the pair's own EvoSuite suite, and keep only pairs both
methods confirm: $994$ of the claimed-equivalent pairs, and $159$ of the $852$ claimed non-equivalent ones, of which $127$ are type-composable. We apply the same verification to BigCloneBench~\citep{svajlenko2014bigclonebench} and find
that most sampled labeled-positive pairs diverge under execution, so we do not
use it. Each level is balanced between positive and negative compositions: 86,
82, 84, and 166 tasks at $N=1,2,3,10$ respectively.

We use a single-turn generation scaffold without tool access, matching the other
single-call domain. The prompt gives both composite programs and asks the model
to decide whether they are semantically equivalent, responding with exactly one
line: \texttt{VERDICT: 1} (equivalent) or \texttt{VERDICT: 0} (not equivalent).
A run ends on model completion or at an 8{,}096-token budget, larger than the
4{,}096 used for complexity analysis because reasoning models can exhaust a
smaller budget before emitting a verdict; requests are retried up to eight times
with exponential backoff on rate limits, server errors, and transient network
failures. The same scaffold and prompt are used for all models. We extract the
verdict by matching \texttt{VERDICT: 0/1}; runs that exhaust the budget without a
parseable verdict are scored incorrect rather than excluded, since they
concentrate on the hardest tasks and excluding them would inflate accuracy.
Kimi-K3 retains 161 of the 166 $N{=}10$ tasks after excluding unresolved
API/runner failures, as in proof completion.

\section{RFT Experimental Setup}
\label{app:rft-hyperparameters}
We use the generated Clone equivalence and Multi-bug repair data to fine-tune Qwen3.5-9B with rejection-sampling fine-tuning (RFT); we do not use all four datasets due to time and cost constraints. The curriculum comprises Base, RFT-L1, RFT-L1+L2, and RFT-L1+L2+L3. Each stage starts from the preceding checkpoint and samples eight agent trajectories per task from the cumulative level pool. At most one successful trajectory per task is retained from the accumulated rollouts. We use one epoch of BF16 LoRA fine-tuning per stage, with rank 16, learning rate $10^{-4}$, and seed 42.

Generation and evaluation use mini-swe-agent with a vLLM backend, temperature $0.7$, top-$p$ $0.95$, a 65,536-token context, and at most 2,048 output tokens per response. Each attempt has a 60-step, 1,800-second agent budget. Multi-bug repair uses 161 training, 53 validation, and 54 reserved test tasks, with disjoint repositories across splits; we report results on the validation split. Candidate repairs are replayed and tested in a fresh sandbox without reference solutions, while Clone equivalence uses label matching. Every checkpoint receives eight evaluation attempts per task; the reported metric is per-attempt success.

\section{Clone equivalence accuracy by class}
\label{app:clone-class}

\Cref{tab:clone-class} splits the accuracies of \cref{tab:results} into positive
(\textsc{eq}) and negative (\textsc{neq}) compositions. Negative accuracy stays
near ceiling at every level for both frontier models, while positive accuracy
falls steeply with depth.

\begin{table}[h]
\centering
\small
\begin{tabular}{@{}llcccc@{}}
\toprule
Model & Class & $N{=}1$ & $N{=}2$ & $N{=}3$ & $N{=}10$ \\
\midrule
{GPT-5.6-Sol}
 & \textsc{eq}  & 97.7 & 80.5 & 78.6 & 45.8 \\
 & \textsc{neq} & 100.0 & 100.0 & 100.0 & 98.8 \\
\addlinespace
{Kimi-K3}
 & \textsc{eq}  & 97.7 & 82.9 & 85.7 & 51.3 \\
 & \textsc{neq} & 93.0 & 92.7 & 95.2 & 94.0 \\
\bottomrule
\end{tabular}
\caption{Clone equivalence accuracy (\%) by composition level, split by class. A positive composition chains $N$ equivalent pairs; a negative one injects a single non-equivalent pair at a random position.}
\label{tab:clone-class}
\end{table}

\end{document}